\documentclass[10pt,twocolumn,letterpaper]{article}

\usepackage[pagenumbers]{cvpr} 

\usepackage{graphicx}
\usepackage{amsmath}
\usepackage{amssymb}
\usepackage{booktabs}

\usepackage[pagebackref,breaklinks,colorlinks]{hyperref}

\usepackage[capitalize]{cleveref}
\crefname{section}{Sec.}{Secs.}
\Crefname{section}{Section}{Sections}
\Crefname{table}{Table}{Tables}
\crefname{table}{Tab.}{Tabs.}

\def\cvprPaperID{*****} 
\def\confName{CVPR}
\def\confYear{2023}

\begin{document}

\title{C2F2NeUS: Cascade Cost Frustum Fusion for High Fidelity and Generalizable Neural Surface Reconstruction}

\author{Luoyuan Xu$^{1}$, Tao Guan$^{1}$, Yuesong Wang$^{1}$\thanks{Corresponding author.  Contact him at yuesongwang@hust.edu.cn}, Wenkai Liu$^{1}$, Zhaojie Zeng$^{1}$, Junle Wang$^{2}$, Wei Yang$^{1}$
\\
$^{1}$ School of Computer Science and Technology, Huazhong University of Science and Technology \\
$^{2}$ Tencent
\\
{\tt\small \{xu\_luoyuan, qd\_gt, yuesongwang, wenkai\_liu, zhaojiezeng, weiyangcs\}@hust.edu.cn, wangjunle@gmail.com}
}


\maketitle

\begin{abstract}
There is an emerging effort to combine the two popular 3D frameworks using Multi-View Stereo (MVS) and Neural Implicit Surfaces (NIS) with a specific focus on the few-shot / sparse view setting. In this paper, we introduce a novel integration scheme that combines the multi-view stereo with neural signed distance function representations, which potentially overcomes the limitations of both methods. MVS uses per-view depth estimation and cross-view fusion to generate accurate surfaces, while NIS relies on a common coordinate volume. Based on this strategy, we propose to construct per-view cost frustum for finer geometry estimation, and then fuse cross-view frustums and estimate the implicit signed distance functions to tackle artifacts that are due to noise and holes in the produced surface reconstruction. We further apply a cascade frustum fusion strategy to effectively captures global-local information and structural consistency. Finally, we apply cascade sampling and a pseudo-geometric loss to foster stronger integration between the two architectures. Extensive experiments demonstrate that our method reconstructs robust surfaces and outperforms existing state-of-the-art methods.
\end{abstract}

\section{Introduction}


Reconstructing 3D structures from a set of images is a fundamental task in computer vision, with widespread applications in fields such as architectural preservation, virtual/augmented reality, and digital twins.
Multi-view stereo (MVS) is a widely-used technique for addressing this task, exemplified by MVSNet~\cite{yao2018mvsnet} and its successors~\cite{yang2020cost, wang2020mesh, wang2021deepfusion, wang2021patchmatchnet, xu2021exploiting}. These methods construct 3D cost volumes based on the camera frustum, rather than regular euclidean space, to achieve precise depth map estimation. However, these methods typically require post-processing steps, such as depth map filtering, fusion, and mesh reconstruction, to reconstruct the 3D surface of the scene, and can not well handle noises, textureless regions, and holes.

The implicit scene representation approaches, e.g., Neural Radiance Fields (NeRF)~\cite{Mildenhall2020NeRFRS} and its peer Neural Signed Distance Function~\cite{takikawa2021nglod, sitzmann2020metasdf, Wang2021NeuSLN}, achieves remarkable results in view synthesis and scene reconstruction. The implicit surface reconstruction approaches typically employ Multi-layer Perceptrons (MLPs) to implicitly fit a volume field. We then can extract scene geometry and render views from the implicit volume field. These approaches usually require a large number of images from different viewpoints and adopt a per-scene optimization strategy, which means they are not generalizable to unknown scenes.

\begin{figure}
	\centering
	\includegraphics[width=0.9\linewidth]{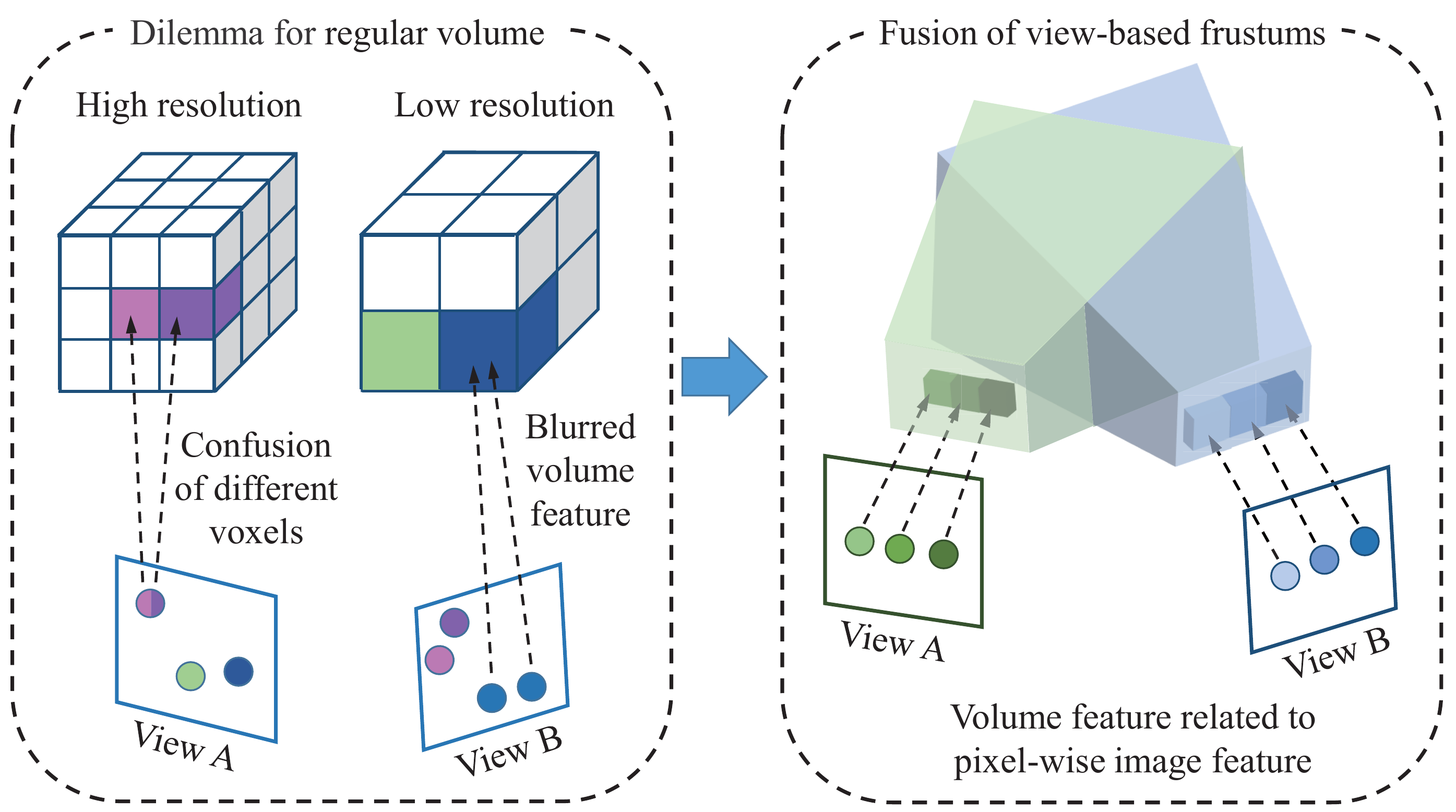}
	\caption{A regular volume doesn't simultaneously fit all cameras well, and can easily run into a dilemma when choosing the resolution. In a low-resolution volume, a single voxel may cover multiple pixels of an image, resulting in blurred volume features, such as the blue pixel in view B. In a high-resolution volume, multiple voxels may cover only a single pixel of an image, causing confusion of different voxels, such as the pink-purple pixels in view A. Instead, frustum volume is view-dependent and extracts pixel-level image features. Hence, we build the cost frustum for each view and adopt the fusion of per-view frustum so as to fit each view.
}
	\vspace{-10pt}
	\label{figure_pipe0}
    
\end{figure}
%

%
There is an emerging effort~\cite{Chen2021MVSNeRFFG, Chang2022RCMVSNetUM, Zhang2021LearningSD, Johari2021GeoNeRFGN} to merge the two technical paths. MVSNeRF~\cite{Chen2021MVSNeRFFG} combines NeRF~\cite{Mildenhall2020NeRFRS} with MVSNet~\cite{yao2018mvsnet} for generalizable view synthesis. RC-MVSNet~\cite{Chang2022RCMVSNetUM} utilizes NeRF's neural volume rendering to handle view-dependent effects and occlusions. The most related to our approach is the SparseNeUS~\cite{long2022sparseneus} for generalizable surface reconstruction method for sparse views. It builds a regular euclidean volume (i.e., cube) to encode geometric information by aggregating 2D feature maps of multiple images. The features sampled from it and the corresponding positions are used to estimate the signed distance function (SDF). However, a regular volume doesn't fit a camera's view naturally, which can be better modeled as view frustum. More specifically, as illustrated in Fig.~\ref{figure_pipe0}, a volume with a higher resolution costs more memory and collect redundant image features, while a coarser volume causes quality degradation. Instead, we propose to build the cost frustum for each view and this strategy has been proven to be effective on MVSNet~\cite{yao2018mvsnet} and its successors.

%

In this paper, we propose a novel integration scheme that combines MVS with neural implicit surface reconstruction. 
To encode the global and local geometric information of the scene, we adopt the cascade architecture of CasMVSNet~\cite{gu2020cascade}, which is a volume pyramid.  Specifically, we first construct a volume on the camera frustum and then convert it into a cascade geometric frustum.
As shown in Fig.~\ref{figure_pipe0}, to fit each camera's view well, we build a cascade frustum for every view and then fuse them using a proposed cross-view and cross-level fusion strategy that effectively captures global-local information and structural consistency.
By combining the 3D position, fused feature, and view direction, we estimate the SDF and render colors using volume rendering~\cite{Wang2021NeuSLN}. Moreover, we utilize the intermediate information output by MVS part to apply cascade sampling and a pseudo-geometric loss, which further improves the quality of the reconstructed surface. Our experiments on the DTU~\cite{aanaes2016large} and BlendedMVS~\cite{yao2020blendedmvs} datasets demonstrate the effectiveness and generalization ability of our proposed method, surpassing existing state-of-the-art generalization surface reconstruction techniques.

Our approach makes the following contributions:
\vspace{-4pt}
\begin{itemize}
  \item{We introduce a novel exploration approach that integrates MVS and implicit surface reconstruction architectures for end-to-end generalizable surface reconstruction from sparse views.}
  \item{We propose a cross-view and cross-level fusion strategy to effectively fuse features from multiple views and levels.}
  \item{We further utilize information from the MVS part to apply cascade sampling and a pseudo-geometric loss to the neural surface part, promoting better integration between the two architectures.}

\end{itemize}


\section{Related Work}

\paragraph{Neural Surface Reconstruction}
Neural implicit representations enable the representation of 3D geometries as continuous functions that are computable at arbitrary spatial locations. Due to the ability to represent complex and detailed shapes in a compact and efficient manner, these representations show significant potential in tasks such as 3D reconstruction~\cite{jiang2020sdfdiff,darmon2022improving, yariv2020multiview, niemeyer2020differentiable, kellnhofer2021neural, Wang2021NeuSLN, Yariv2021VolumeRO, yu2022monosdf, oechsle2021unisurf}, shape representation~\cite{atzmon2020sal, gropp2020implicit, mescheder2019occupancy, park2019deepsdf}, and novel view synthesis~\cite{Mildenhall2020NeRFRS, liu2020neural, saito2019pifu, trevithick2021grf}.


To avoid relying on ground-truth 3D geometric information, many of these methods employ 2D images as supervision through classical rendering techniques, such as surface rendering and volume rendering. While some methods~\cite{niemeyer2020differentiable, kellnhofer2021neural, liu2020dist, yariv2020multiview} reconstruct the surface and render 2D images using surface rendering, they often require accurate object masks, which can be challenging to obtain in practical scenarios. As NeRF~\cite{Mildenhall2020NeRFRS} successfully integrates implicit neural functions and volume rendering and generates photo-realistic novel views, some methods~\cite{darmon2022improving,oechsle2021unisurf, Wang2021NeuSLN, Yariv2021VolumeRO} incorporate SDF into neural volume rendering to achieve surface reconstruction without additional masks.  Despite these advancements, further improvement in surface quality is achieved by introducing additional geometric priors~\cite{Zhang2021LearningSD, fu2022geo}. However, these methods require a large number of dense images, and it is difficult to generalize to unknown scenes, which restricts the deployment at the level of these methods. 

Some methods~\cite{Chen2021MVSNeRFFG, chibane2021stereo, liu2022neural, wang2021ibrnet, yu2021pixelnerf} generate novel views in unknown scenarios in a generalization manner. These methods construct radiative neural fields on sparse views, and can inference on unknown scenarios without any fine-tuning after training in multiple known scenarios. Moreover, some other methods~\cite{niemeyer2022regnerf, jain2021putting, deng2022depth} synthesize novel views on a single scene with sparse views. However, these methods are difficult to generate high-quality geometries.

To overcome these deficiencies, SparseNeUS~\cite{long2022sparseneus} provides a preliminary solution by encoding geometric information using a regular euclidean volume, VolRecon~\cite{ren2023volrecon} introduces multi-view image features through the view transformer to advance this scheme, ReTR~\cite{liang2023rethinking} uses hybrid extractor to obtain multi-level euclidean volume and then utilize reconstruction transformer to improve the performance. However, these methods are challenging to achieve high-quality reconstructions due to the regular volume doesn't fit a camera's view naturally. Moreover, VolRecon~\cite{ren2023volrecon} and  ReTR~\cite{liang2023rethinking} additionally introduce ground truth depth labels, which are usually expensive to obtain.

\paragraph{Multi-view Stereo}
With the rapid advancements in deep learning techniques, MVS methods~\cite{yao2018mvsnet, wang2021patchmatchnet, Wei2021AARMVSNetAA, luo2020attention, cheng2020deep, luo2019p, wang2023adaptive, wang2020mesh} based on depth map fusion have shown remarkable performance on various benchmarks~\cite{aanaes2016large, yao2020blendedmvs}. The pioneering MVSNet~\cite{yao2018mvsnet} architecture constructs a 3D cost volume by leveraging differentiable homography warping operations, and generates the depth map through cost volume regularization. The key to success lies in its utilization of camera frustums instead of regular euclidean spaces for constructing 3D cost volumes. 
Some attempts~\cite{yang2020cost, gu2020cascade} progressively optimize the depth map by refining the camera frustum in a coarse-to-fine manner. Some attempts~\cite{ding2022transmvsnet, zhu2021multi, wang2022mvster, liao2022wt} introduce transformers to improve reconstruction performance. Some other attempts~\cite{xu2021exploiting, xu2022self, ding2022kd, xu2021self, yang2021self, xu2021digging, Chang2022RCMVSNetUM} train networks in an unsupervised manner.
However, these methods require a series of post-processing operations, such as depth map filtering, depth map fusion, and mesh reconstruction, to reconstruct the 3D structure of the scene, and can not well handle noises, textureless regions, and holes.

\begin{figure*}
	\centering
	\includegraphics[width=1.\linewidth]{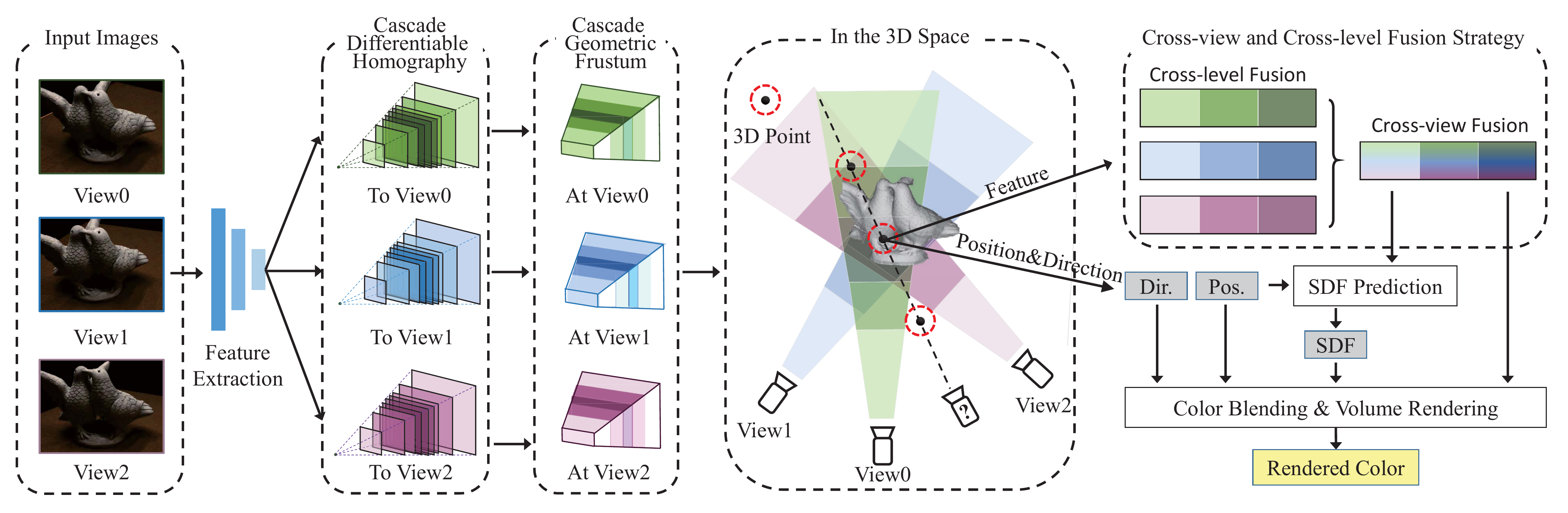}
	\caption{The overview of C2F2NeUS. We first construct the cascade geometric frustum of each view to capture global and local geometric information about the scene. Then we apply a cross-view and cross-level fusion strategy to effectively fuse features from multiple views and levels. Finally, the positions of 3D points and their corresponding fused features are fed to an SDF prediction network which is trained by rendered colors from volume rendering.}
	\vspace{-10pt}
	\label{figure_pipeline}
    
\end{figure*}

\paragraph{The Integration of MVS and Neural Implicit Scene Representation}
The integration of MVS and neural implicit scene representation generates significant interest among researchers, leading to several recent explorations~\cite{Chen2021MVSNeRFFG, Chang2022RCMVSNetUM, Zhang2021LearningSD, Johari2021GeoNeRFGN}. MVSNeRF~\cite{Chen2021MVSNeRFFG} constructs a cost volume to enable geometry-aware scene reasoning. It then uses volume rendering~\cite{Mildenhall2020NeRFRS} in combination with position, view direction, and volume features to perform neural radiation field reconstruction and achieve generalizable view synthesis. 
RC-MVSNet~\cite{Chang2022RCMVSNetUM} adds an independent cost volume for volume rendering, allowing the network to learn how to handle view-dependent effects and occlusions and improve the quality of depth maps. MVSDF~\cite{Zhang2021LearningSD} leverages the geometry and feature consistency of Vis-MVSNet\cite{zhang2020visibility} to optimize the SDF, resulting in more robust geometry estimation. 

However, MVSNeRF~\cite{Chen2021MVSNeRFFG} is difficult to generate high-quality surfaces, RC-MVSNet~\cite{Chang2022RCMVSNetUM} requires cumbersome post-processing steps to obtain surfaces, and MVSDF~\cite{Zhang2021LearningSD} cannot generalize to unknown scenes and requires dense images. Our method, on the other hand, differs significantly as it focuses on achieving high fidelity and generalizable surface reconstruction for sparse views.

\section{Method}

In this section, we explain the detailed structure of our proposed C2F2NeUS, which is a novel integration scheme that better combines MVS with neural implicit surface representation. With this integration, C2F2NeUS achieves high fidelity and generalizable surface reconstruction for sparse views in an end-to-end manner. As illustrated in Fig.~\ref{figure_pipeline}, by fusing the view-dependent frustums in the MVS part, we obtain more accurate geometric features which are sent to the neural implicit surface part to predict SDF and extract surfaces. 

Specifically, we first construct a view-dependent cascade geometric frustum for each view to encode geometric information of the scene and fully exploit the advantage of MVS(Sec.~\ref{method_vol_gen}). 
For a given set of 3D coordinates, we then sample and fuse the feature from these frustums by using the proposed cross-view and cross-level fusion strategy (Sec.~\ref{method_fea_fus}). This strategy can effectively capture global-local information and structural consistency. Next, we introduce how to predict SDF and render color from the fused feature (Sec.~\ref{method_sdf_color}). And the SDF prediction network generates an SDF field which is used for surface reconstruction, this representation leverages the smooth and complete geometry of SDF. To train the SDF prediction network in an unsupervised manner, we render color via volume rendering. Finally, we introduce the training loss of our end-to-end framework (Sec.~\ref{method_loss_fun}). 

\subsection{Cascade Geometric Frustum Generation}
\label{method_vol_gen}

To make the implicit neural surface reconstruction generalizable and capture the scene information more accurately, we follow the volume pyramid of CasMVSNet~\cite{gu2020cascade} and encode the global and local geometric information of the scene by building a cascade geometric frustum.
Unlike SparseNeUS~\cite{long2022sparseneus}, which utilizes a regular euclidean volume, we construct the volume from the perspective MVS. In MVS, the reference view is most important, and other source views contribute to the depth estimation for the reference view. Since a single view-dependent frustum cannot describe the complete scene, we create a frustum for each image and treat the image as the reference view and other images as source views. Besides, we estimate the corresponding depth maps from each cost frustum to construct the cascade geometric frustums.

To accomplish this, we first extract feature maps $ \{F_{i}\}_{i=0}^{N-1} $  using a 2D feature extraction network for $N$ images $ \{I_{i}\}_{i=0}^{N-1} $ of the scene. With the corresponding camera parameters $ \{K_{i}, R_{i}, T_{i}\}_{i=0}^{N-1} $ of each image, we then build a 3D cost frustum $C \in \mathbb{R}^{c \times d \times h \times w}$ for the reference camera via differential homography warping operations, where $c$, $d$, $h$, $w$ are dimensions of feature, number of depth samples, height, width respectively. In our implementation, we construct a 3D cost frustum $ \{C_{i}\}_{i=0}^{N-1} $ for each image as the reference view and use the remaining images as source views. 
The 3D cost frustums $ C_{i} $ are then regularized by 3D CNN $\Psi_{1}$ to obtain the intermediate volumes $V_{i} \in \mathbb{R}^{c \times d \times h \times w}$. On the one hand, the intermediate volumes $V_{i}$ are used to estimate the probability volumes $ P_{i} \in \mathbb{R}^{1 \times d \times h \times w} $ which are used to regress the depth maps $ D_{i} \in \mathbb{R}^{1 \times 1 \times h \times w} $ of the current reference views $ I_{i} $. On the other hand,  the intermediate volumes $V_{i}$ are further regularized by a new 3D CNN $\Psi_{2}$ to obtain the geometric frustums $ G_{i} \in \mathbb{R}^{c \times d \times h \times w} $.
\begin{equation}
\label{equ_3dcnn}
V_{i}, P_{i}, D_{i}=\Psi_{1}(C_{i}), \qquad G_{i}=\Psi_{2}(V_{i}).
\end{equation}
The depth maps $ D_{i} $ are used to redefine the depth hypothesis and construct the cascade 3D cost frustums, ultimately constructing cascade geometric frustums $ \{G_{i}^{j}\}_{i=0,...,N-1}^{j=0,...,L-1} $, where $L$ is the cascade level number.

\subsection{Cross-view and Cross-level Fusion Strategy}
\label{method_fea_fus}

The cost frustum of each view and level has different importance. Intuitively, regions with relatively smaller angles w.r.t. the viewpoint in finer frustums are more crucial. Therefore, we estimate the weight for each frustum to represent importance, similar to Vis-MVSNet\cite{zhang2020visibility}. Another problem in the fusion process is that points near the surface can sample features from all three pyramid levels, but points far away from the surface can only extract features from the coarse level. Straightforward fusion strategies, such as adding features from all views and levels, will confuse features of different levels, while directly concatenating all features together make it difficult to deal with an arbitrary number of input views. Consequentially, we propose a cross-view and cross-level fusion strategy that treats each view and level differently. This fusion strategy effectively captures the spatial and structural information of the scene and produces a more precise surface.

We introduce an adaptive weight $A_{i}^{j} \in \mathbb{R}^{1 \times d \times h \times w}$ for each geometric frustum $ G_{i}^{j} $, which is normalized using the sigmoid function. Therefore, we can rewrite Equ.~\ref{equ_3dcnn} as
\begin{equation}
\label{equ_3dcnn2}
V_{i}, P_{i}, D_{i}=\Psi_{1}(C_{i}), \qquad G_{i}, A_{i}=\Psi_{2}(V_{i}).
\end{equation}
To integrate both the global information from coarser frustums and the local information of finer frustums, we concatenate features at different levels and sum the features at different viewpoints according to their weights. Specifically, we sample the corresponding features $ g_{i}^{j} = G_{i}^{j}(p) 
 \in \mathbb{R}^{1 \times c}$ and weights $ a_{i}^{j} = A_{i}^{j}(p)\in \mathbb{R}^{1 \times 1} $ of a given 3D position $p \in \mathbb{R}^{1 \times 3}$ from all frustums using bilinear interpolation. 
 Then, we concatenate features and sum the weights of different levels $j=0,...L-1$ for each viewpoint $I_{i}$, and obtain new features $ g_{i}^{L} = \text{cat}(\{g_{i}^{j}\}), $ and new weights $ a_{i}^{L}(p) = \text{sum}(\{a_{i}^{j}\})$, respectively, where $ g_{i}^{L}(p)  \in \mathbb{R}^{ 1 \times Lc}$ and $ a_{i}^{L}(p) \in \mathbb{R}^{1 \times 1}$.
 Finally, we fuse the concatenated features $g_{i}^{L}$ from different viewpoints $ \{I_{i}\}_{i=0}^{N-1} $ based on their respective weights $a_{i}^{L}$. The final geometric feature of the given 3D position p is defined as
$f_{geo} = {\Sigma_{i=0}^{N-1} a_{i}^{L} \cdot g_{i}^{L}}/{\Sigma_{i=0}^{N-1} a_{i}^{L}}$, where $f_{geo}  \in \mathbb{R}^{1 \times Lc}$.

\subsection{SDF Prediction and Volume Rendering}
\label{method_sdf_color}
We would like to exploit the advantage of neural implicit surface reconstruction, i.e., the surface extracted from a neural SDF network is usually very smooth and consistent.

\paragraph{SDF Prediction.} Given an SDF prediction network $\Phi$ consisting of MLP and an arbitrary 3D position $p$ with its corresponding geometric feature $f_{geo}$, we first encode the position $p$ using position encoding $\gamma (\cdot)$. We then use the encoded position and geometric feature $f_{geo}$ as input to the SDF prediction network $\Phi$ to predict the SDF $s(p)$ of 3D position $p$. Our SDF prediction operation is defined as:
\begin{equation}
\label{equ_sdf_pred}
s(p)=\Phi \big ( \gamma (p), f_{geo} \big) .
\end{equation}

\paragraph{Blending Weights. } Similar to IBRNet~\cite{wang2021ibrnet}, we use blending weights to estimate color of a 3D position $p$ and view direction $r$. We extract 2D color feature maps from $N$ input images via a new feature extract network. For a given 3D position $p$ with its corresponding geometric feature $f_{geo}$ and view direction $r$, we project $p$ onto $N$ input views and extract corresponding color features $f_{i}^{col}$ from color feature maps using bilinear interpolation. We then compute the mean $u$ and variance $v$ of the sampled color features $f_{i}^{col}$ for different views to capture cross-image information and concatenate each feature $f_{i}^{col}$ with $u$ and $v$. A small shared MLP $\Gamma$ is used to process the concatenated features and generate new features $f_{i}^{col2}$ that contain color information. We also compute the direction difference $\Delta r=r-r_{i}$ between the view direction $r$ and each input image's viewpoint $r_{i}$. The color features $f_{i}^{col2}$, direction differences $\Delta r$, and geometric features $f_{geo}$ are fed into a new MLP network $\Gamma_{col}$ for generating blending weights $w_{i}(p)$.
\begin{equation}
\label{equ_ble_wei}
w_{i}(p) = \Gamma_{col}(\Gamma(f_{i}^{col}, u, v), \Delta r, f_{geo}).
\end{equation}
Finally, we use the softmax operator to normalize blending weights $\{w_{i}(p)\}_{i=0}^{N-1}$.

\paragraph{Volume Rendering.} As there is no ground-truth 3D geometry, to supervise the SDF prediction network, we render the color of the query ray and calculate its consistency with the ground-truth color. Specifically, we perform the ray point sampling, where each sampled position $p$ and viewpoint $d$ are used to predict the corresponding SDF $s(p)$ and blending weights $w_{i}(p)$. We then project the position $p$ onto $N$ input images to extract their respective colors $c_{i}(p)$ and compute the color $c(p)$ of position $p$ as a weighted sum of the sampled color $c_{i}(p)$ and blending weights $w_{i}(p)$. Next, we apply volume rendering as in NeUS~\cite{Wang2021NeuSLN} to render the color of the ray by aggregating the SDF and color of each position $p$ along the ray. The rendered color is compared to the ground-truth color for calculating the consistency loss.

\paragraph{Cascade Sampling and Pseudo-depth Generation.}

To further leverage the benefits of MVS and enhance the quality of the extracted surfaces, we incorporate cascade sampling on the frustum and a pseudo-geometric loss, which enforce a stronger integration between MVS and neural implicit surface.
%
In this work, we apply an adaptive sampling strategy using the intermediate probability volume $P_{i}$ of the cascade frustum generation network. Specifically, we take the query image as the reference view, and other input images as the source views, and send them to the cascade frustum generation network to obtain the depth maps $D_{que}^{j}$ and probability volumes $P_{que}^{j}$ at different levels. We use the probability volume $P_{que}^{j=0}$ of the coarsest layer for cascade sampling. We then compute the mean $\alpha$ and standard $\beta$ deviation of the probability volume $P_{que}^{j=0}$ along the depth channel. The adaptive sample ranges $[t_{n}, t_{f}]$ are defined as follows:
\begin{equation}
\label{equ_mix_sam}
[t_{n}, t_{f}] = [\alpha-\beta, \alpha+\beta].
\end{equation}
With the high-resolution depth maps ${D_{que}^{j=L-1}, D_{i}^{j=L-1}}$ of the query image and other input images, we compute the geometric consistency to obtain the respective effective masks. The masked depth maps can be considered pseudo-depth labels, which use to supervise the SDF prediction network. Specifically, the masked depth map of the query image is used to compute a pseudo-depth consistency loss. Further, we fuse the masked depth maps of different images into point clouds, which are used to directly supervise the SDF prediction network.

\subsection{Loss Function}
\label{method_loss_fun}

Ground-truth 3D geometric labels are difficult to obtain, to address this issue, our framework employs an unsupervised learning approach. Specifically, we introduce a training loss $\mathcal{L}_{total}$ as a combination of two unsupervised losses for training the depth map and SDF, respectively.
\begin{equation}
\label{equ_loss_dep}
\mathcal{L}_{total} = \mathcal{L}_{dep} + \mathcal{L}_{sdf}.
\end{equation}

For the loss $\mathcal{L}_{dep}$, we supervise the intermediate depth map $D_{i}^{j}$ of cascade geometric frustum network by utilizing several losses of SMU-MVSNet~\cite{xu2021exploiting}.

For the loss $\mathcal{L}_{sdf}$, we adopt the same loss item of SparseNeUS~\cite{long2022sparseneus}, i.e. the color consistency loss $\mathcal{L}_{cc}$, the Eikonal term $\mathcal{L}_{eik}$, the sparseness regularization term $\mathcal{L}_{spa}$. Moreover, we introduce a geometry-based loss derived from pseudo-depth, which can provide reliable guidance without relying on expensive ground-truth geometry labels. The overall loss $\mathcal{L}_{sdf}$ is defined as follows:
\begin{equation}
\label{equ_loss_sdf}
\mathcal{L}_{sdf} = \mathcal{L}_{cc} + \mathcal{L}_{eik} +\mathcal{L}_{spa} + \mathcal{L}_{pdc} + \mathcal{L}_{pgs}.
\end{equation}

The pseudo-depth consistency loss $\mathcal{L}_{pdc}$ is an L1 distance between the rendered depth and the pseudo-depth label. $\mathcal{L}_{pdc}$ is defined as:
\begin{equation}
\label{equ_loss_sdf_10pdc}
\mathcal{L}_{pdc} = \frac{1}{X} \sum\limits_{x=0}\limits^{X-1}|d-\hat d|_{1},
\end{equation}
\noindent where $d$ and $\hat d$ are the rendered depth and ground-truth depth respectively.
$\mathcal{L}_{pgs}$ is the pseudo-geometry SDF loss~\cite{fu2022geo}. The SDF values of the pseudo point cloud are zeroes. $\mathcal{L}_{pgs}$ is defined as:
\begin{equation}
\label{equ_loss_sdf_11pgs}
\mathcal{L}_{pgs} = \frac{1}{||Q_{2}||} \sum\limits_{q_{2} \in Q_{2}} |s(q_{2})|,
\end{equation}
\noindent where $Q_{2}$ is a set of 3D points randomly selected from the pseudo point clouds.

\section{Experiments}

In this section, we demonstrate the effectiveness of our proposed method. Firstly, we provide a detailed account of our experimental settings, which includes implementation details, datasets, and baselines. Secondly, we present quantitative and qualitative comparisons on two widely used datasets, namely DTU~\cite{aanaes2016large} and BlendedMVS~\cite{yao2020blendedmvs}. Finally, we conduct detailed ablation studies to analyze the contribution of different components of our proposed method.

\begin{figure*}
	\centering
	
	\includegraphics[width=1.\linewidth]{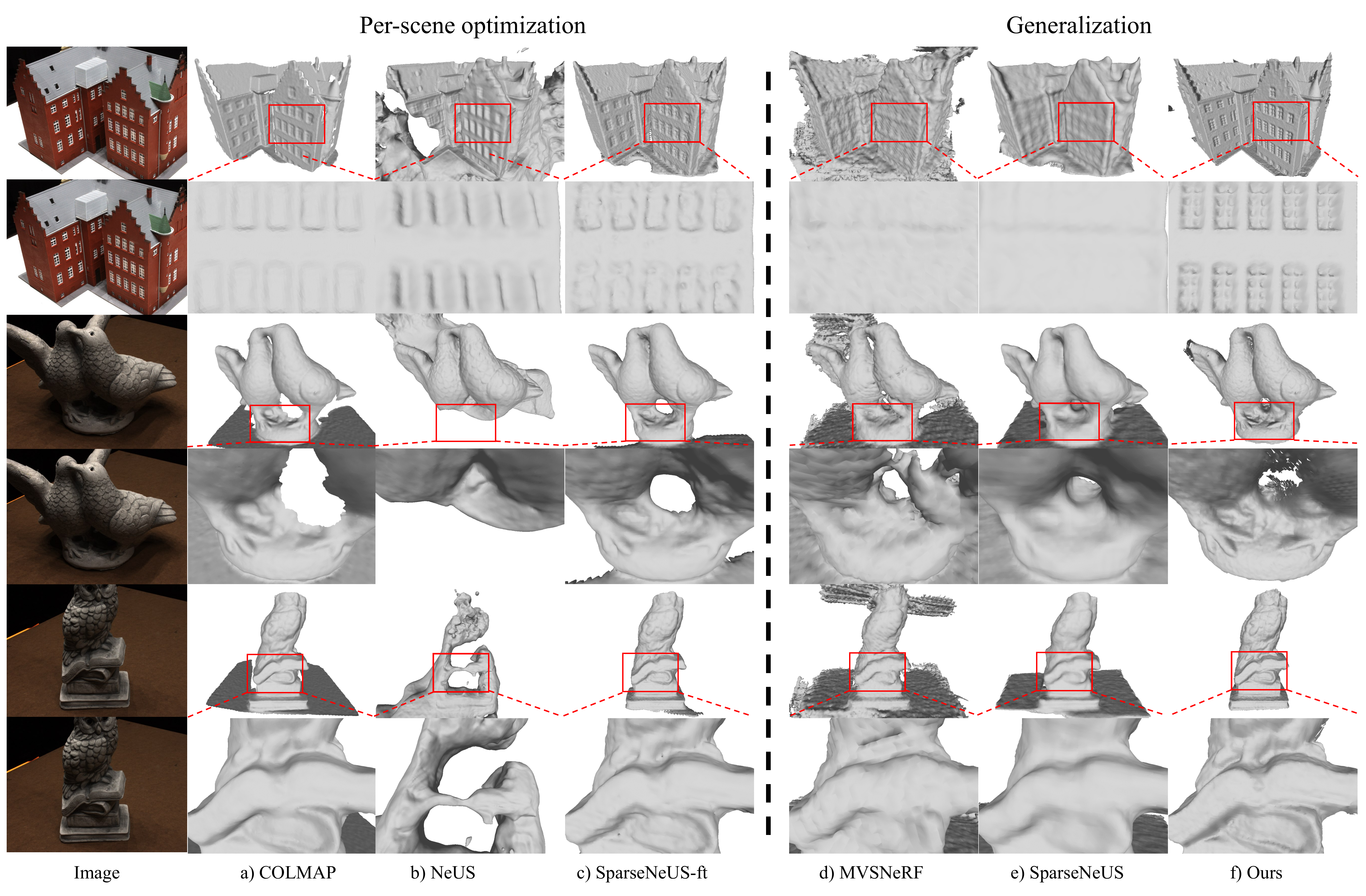}
	\vspace{-20pt}
	\caption{The qualitative comparison of our method with other state-of-the-art methods on DTU. }%
	\vspace{-10pt}
	\label{figure_result_dtu}
\end{figure*}

\begin{table*}

	\begin{center}
 \resizebox{\textwidth}{!}{
		\begin{tabular}{ c| c c c c c c c c c c c c c c c | c}
			\hline
			Scan & 24 & 37 & 40 & 55 & 63 & 65 & 69 & 83 & 97 & 105 & 106 & 110 & 114 & 118 & 122 & \textbf{Mean} \\
			\hline\hline
   COLMAP~\cite{schonberger2016structure} & \textbf{0.90} & 2.89 & 1.63 & 1.08 & 2.18 & 1.94 & 1.61 & 1.30 & 2.34 & 1.28 & 1.10 & 1.42 & 0.76 & 1.17 & 1.14 & 1.52 \\
   \hline
   IDR~\cite{yariv2020multiview} & 4.01 & 6.40 & 3.52 & 1.91 & 3.96 & 2.36 & 4.85 & 1.62 & 6.37 & 5.97 & 1.23 & 4.73 & 0.91 & 1.72 & 1.26 & 3.39 \\
   VolSDF~\cite{Yariv2021VolumeRO} & 4.03 & 4.21 & 6.12 & 0.91 & 8.24 & 1.73 & 2.74 & 1.82 & 5.14 & 3.09 & 2.08 & 4.81 & 0.60 & 3.51 & 2.18 & 3.41 \\
   UNISURF~\cite{oechsle2021unisurf} & 5.08 & 7.18 & 3.96 & 5.30 & 4.61 & 2.24 & 3.94 & 3.14 & 5.63 & 3.40 & 5.09 & 6.38 & 2.98 & 4.05 & 2.81 & 4.39 \\
   NeUS~\cite{Wang2021NeuSLN} & 4.57 & 4.49 & 3.97 & 4.32 & 4.63 & 1.95 & 4.68 & 3.83 & 4.15 & 2.50 & 1.52 & 6.47 & 1.26 & 5.57 & 6.11 & 4.00 \\
   
   IBRNet-ft~\cite{wang2021ibrnet} & 1.67 & 2.97 & 2.26 & 1.56 & 2.52 & 2.30 & 1.50 & 2.05 & 2.02 & 1.73 & 1.66 & 1.63 & 1.17 & 1.84 & 1.61 & 1.90 \\
   
   SparseNeUS-ft~\cite{long2022sparseneus} & 1.29 & \textbf{2.27} & 1.57 & 0.88 & 1.61 & 1.86 & 1.06 & 1.27 & 1.42 & 1.07 & 0.99 & 0.87 & 0.54 & 1.15 & 1.18 & 1.27 \\
   \hline
			PixelNerf~\cite{yu2021pixelnerf} & 5.13 & 8.07 & 5.85 & 4.40 & 7.11 & 4.64 & 5.68 & 6.76 & 9.05 & 6.11 & 3.95 & 5.92 & 6.26 & 6.89 & 6.93 & 6.28 \\
                IBRNet~\cite{wang2021ibrnet} & 2.29 & 3.70 & 2.66 & 1.83 & 3.02 & 2.83 & 1.77 & 2.28 & 2.73 & 1.96 & 1.87 & 2.13 & 1.58 & 2.05 & 2.09 & 2.32 \\
                MVSNeRF~\cite{Chen2021MVSNeRFFG} & 1.96 & 3.27 & 2.54 & 1.93 & 2.57 & 2.71 & 1.82 & 1.72 & 2.29 & 1.75 & 1.72 & 1.47 & 1.29 & 2.09 & 2.26 & 2.09\\
                SparseNeUS~\cite{long2022sparseneus} & 1.68 & 3.06 & 2.25 & 1.10 & 2.37 & 2.18 & 1.28 & 1.47 & 1.80 & 1.23 & 1.19 & 1.17 & 0.75 & 1.56 & 1.55 & 1.64\\
                VolRecon$^{\dag}$~\cite{ren2023volrecon}  & 1.20 & 2.59 & 1.56 & 1.08 & 1.43 & 1.92 & 1.11 & 1.48 & 1.42 & 1.05 & 1.19 & 1.38 & 0.74 & 1.23 & 1.27 & 1.38 \\
                ReTR$^{\dag}$~\cite{liang2023rethinking}  & 1.05 & 2.31 & 1.44 & 0.98 & \textbf{1.18} & \textbf{1.52} & 0.88 & 1.35 & 1.30 & 0.87 & 1.07 & 0.77 & 0.59 & 1.05 & 1.12 & 1.17 \\
                Ours & 1.12 & 2.42 & \textbf{1.40} & \textbf{0.75} & 1.41 & 1.77 & \textbf{0.85} & \textbf{1.16} & \textbf{1.26} & \textbf{0.76} & \textbf{0.91} & \textbf{0.60} & \textbf{0.46} & \textbf{0.88} & \textbf{0.92} & \textbf{1.11} \\
                \hline
		\end{tabular}
  }
	\end{center}
	\vspace{-15pt}
    \caption{The quantitative results of different methods on DTU. $^{\dag}$ indicates supervised by ground truth depth labels}
    \vspace{-15pt}
	\label{table_result_dtu}
\end{table*}

\begin{figure*}
\centering
\includegraphics[width=1.\linewidth]{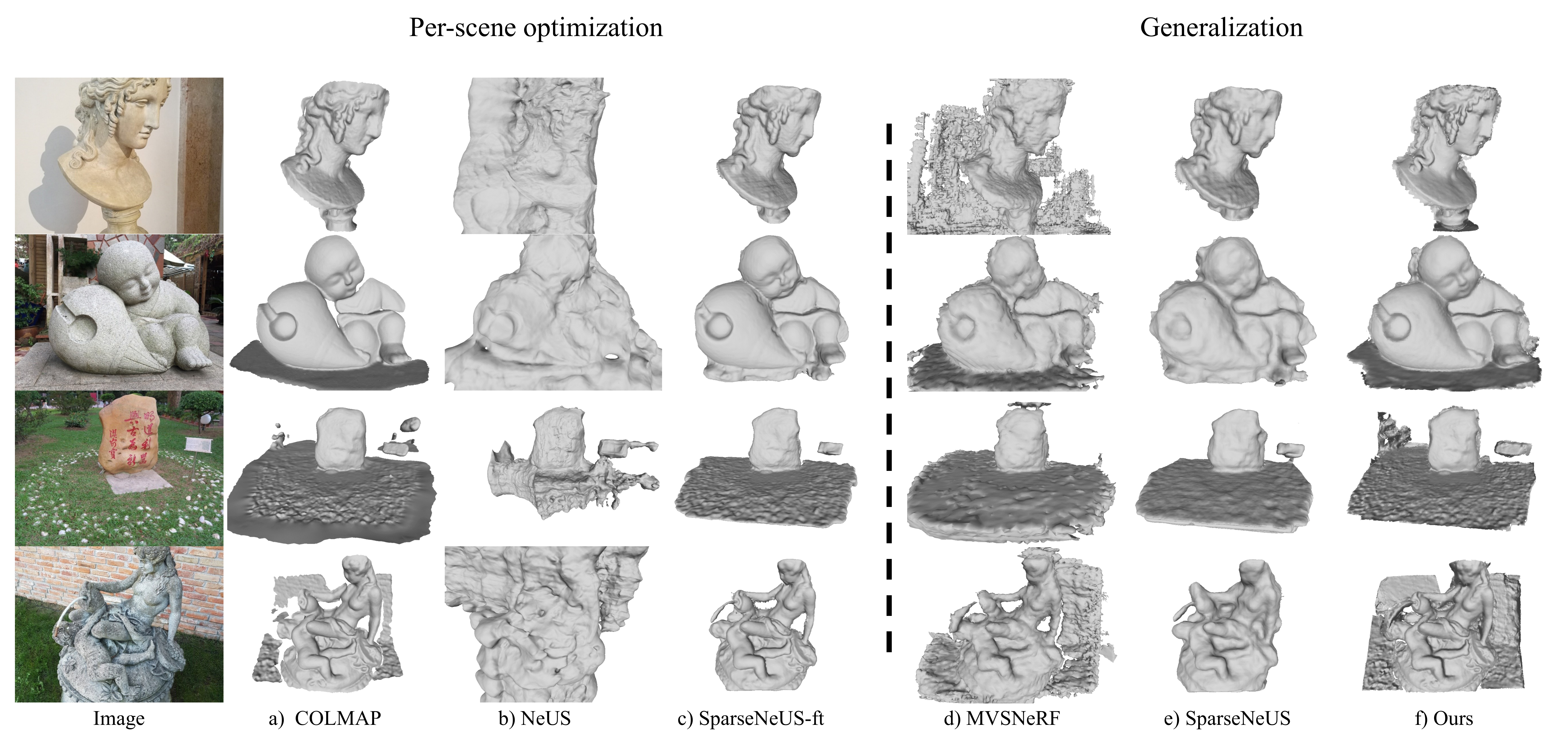}
\vspace{-20pt}
\caption{The qualitative comparison of our method with other state-of-the-art methods on BlendedMVS. We reconstruct the surface with our method without any fine-tuning.}
\vspace{-10pt}
\label{figure_result_bmvs}
    
\end{figure*}

\subsection{Experimental Settings}

\paragraph{Implementation Details.} 
We implement our method in PyTorch. In the cascade geometric frustum generation network, we adopt the same cascade scheme with CasMVSNet~\cite{gu2020cascade}, but make the following changes: we use the 2D feature extraction network consisting of 9 convolutional layers, share the 3D CNN $\Psi$ across levels, and set the dimension of image and volume features to $c=8$.
During training, we take $N=5$ images with a resolution of $640 \times 512$ as input and use an additional image as the query image to supervise the SDF prediction network $\Phi$. The cascade stage number is set to $L=3$. 
We train our end-to-end framework on one A100 GPU with a batch size of 2 for 300k iterations. We set the same learning rate and cosine decay schedule as NeUS~\cite{Wang2021NeuSLN}. The ray number is set to $X=512$, and the sample number on each ray is $Y=N_{coarse}+N_{fine}$, where $N_{coarse}=64$ and $N_{fine}=64$. 
The weights of $\mathcal{L}_{pdc}$ and $\mathcal{L}_{pgs}$  in Equ.~\ref{equ_loss_sdf} are $0,0$ before 10k iterations and $0.05,1$  after that.

\paragraph{Datasets.}
The DTU dataset~\cite{aanaes2016large} is a well-known indoor multi-view stereo dataset, consisting of 124 scenes captured under 7 distinct lighting conditions. Consistent with prior research~\cite{long2022sparseneus, Wang2021NeuSLN, Zhang2021LearningSD, fu2022geo}, we employ 75 scenes for training and 15 non-overlapping scenes for testing. Each test scene contains two sets of three images offered by SparseNeUS~\cite{long2022sparseneus}. We evaluate our method using three views with a resolution of $1600 \times 1152$. To ensure fairness in evaluation, we adopt the foreground masks provided by IDR~\cite{yariv2020multiview} to assess the performance of our approach on the test set, as in previous studies~\cite{long2022sparseneus, Wang2021NeuSLN, Zhang2021LearningSD, fu2022geo}. To examine the generalization ability of our proposed framework, we conduct a qualitative comparison of our method on the BlendedMVS dataset~\cite{yao2020blendedmvs} without any fine-tuning.

%
%
\subsection{Comparisons on DTU}
We perform surface reconstruction for sparse views (only 3 views) on the DTU dataset~\cite{aanaes2016large} and evaluate the predicted surface against the ground-truth point clouds using the chamfer distance metric. Tab.~\ref{table_result_dtu} and Fig.~\ref{figure_result_dtu} present a summary of the comparison between our method and other existing methods, which demonstrate that our method achieves better performance. It is important to note that our method is solely trained on the training set without any fine-tuning on the test set to assess its generalization capability. Our method surpasses the generalizable version of SparseNeUS~\cite{long2022sparseneus} by $32\%$ and significantly outperforms its fine-tuning variant. Furthermore, our method exhibits superior performance compared to VolRecon~\cite{ren2023volrecon} and ReTR~\cite{liang2023rethinking}
, which are the state-of-the-art generalizable neural implicit reconstruction methods and are supervised with ground-truth deep labels

\subsection{Generalization on BlendedMVS}

To showcase the generalization capabilities of our proposed method, we conduct additional tests on the BlendedMVS dataset~\cite{yao2020blendedmvs} without any fine-tuning. The qualitative comparison between our method and other methods is presented in Fig.~\ref{figure_result_bmvs}. The results indicate that our method exhibits a robust generalization ability and produces a more refined surface when compared to other generalizable neural implicit reconstruction methods.

\subsection{Comparison with Unsupervised MVS}

\begin{figure}
\centering
\includegraphics[width=1.\linewidth]{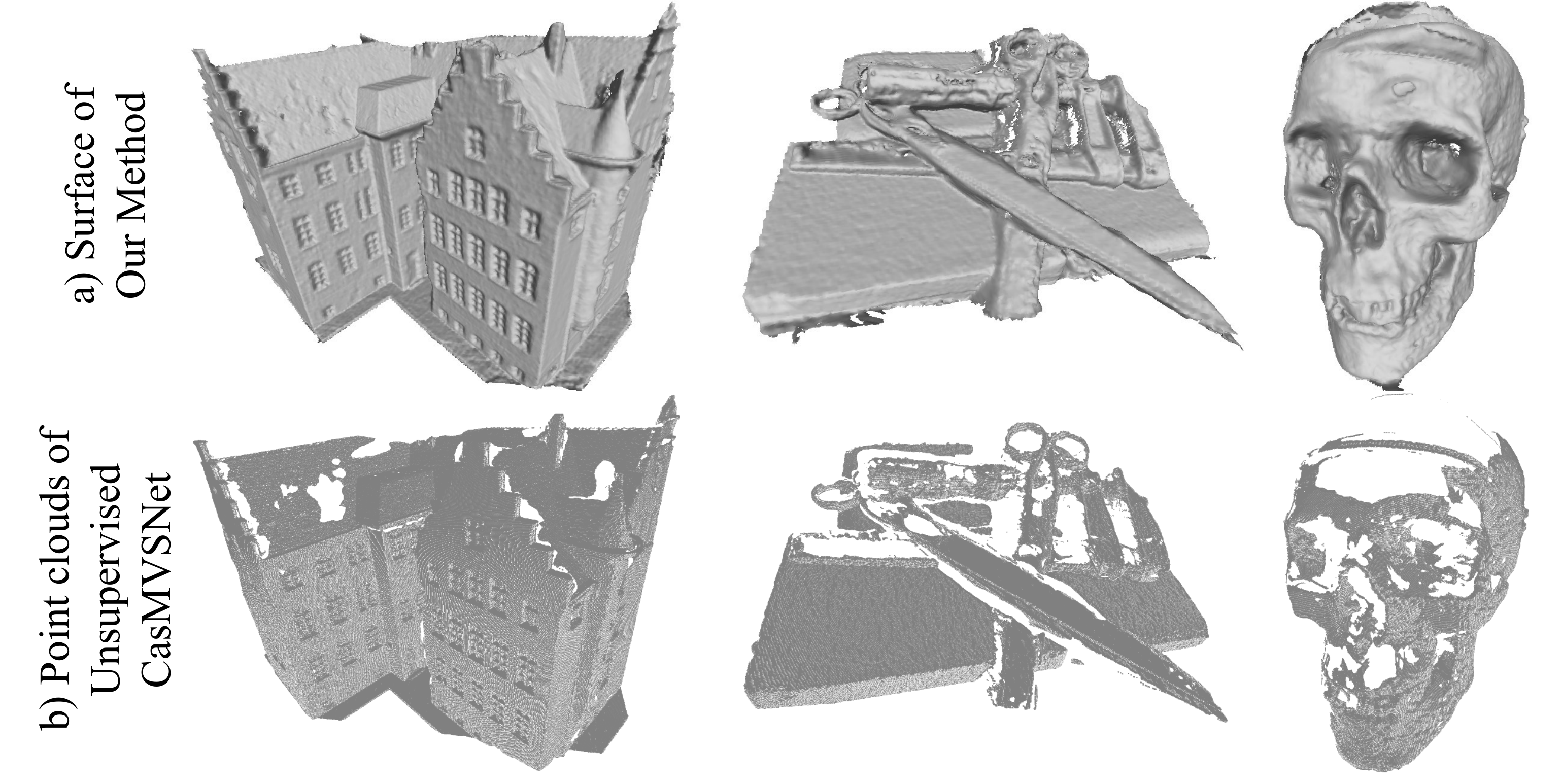}
\caption{The qualitative comparison of our method with MVS. We present the surface reconstructed by our method, and point clouds of unsupervised CasMVSNet.}
\vspace{-15pt}
\label{figure_abl_com_mvs_sdf}
    
\end{figure}


To compare our method with MVS, we retrain CasMVSNet~\cite{gu2020cascade} with the unsupervised loss in Equ.~\ref{equ_loss_dep}. We estimate the depth maps of three views, filter the depth maps using geometric consistency and the masks provided by IDR~\cite{yariv2020multiview}, and fuse them into a point cloud. The qualitative comparison is shown in Fig.~\ref{figure_abl_com_mvs_sdf}, which demonstrates that our method is more robust and reconstructs a more complete surface.

\subsection{Ablation Studies}

\paragraph{Effect of Camera Frustum and Regular Euclidean Space.}

Regular volume doesn't simultaneously fit all cameras well, which leads to blurred features, particularly in sparser scenes. Instead, camera frustum volume can better model. We present a performance comparison between the regular euclidean volume and the camera frustum volume without cascade in Table~\ref{table_frus_eucl}. The comparison results reveal that our method achieves better performance with similar GPU memory. 
\vspace{-2pt}

\begin{table}
    
	\begin{center}
  \resizebox{\linewidth}{!}{
		\begin{tabular}{  c|c c c c}
			\hline
			Method & Inp. Size & Vol. Size & Cham. Dis. & GPU \\
			\hline\hline
			  Euclidean & 800*600 & 96*96*96 & 1.77 & \textbf{2231} M \\
                Euclidean & 800*600 & 192*192*192 & 1.62 & 7073 M \\
                \hline
			Frustum   & 200*144 & 200*144*48 & 1.50 & 2291 M\\
                Frustum   & 400*288 & 400*288*48 & \textbf{1.42} & 4621 M\\
			\hline
		\end{tabular}
  }
	\end{center}
	\vspace{-10pt}
    \caption{Effect of camera frustum and regular euclidean space. We evaluate the two methods without cascade under different image sizes and volume sizes. Our method achieves a better performance.}
    \vspace{-5pt}
	\label{table_frus_eucl}
\end{table}

\paragraph{Effect of Different Components.}

In this experiment, we present the results of different components to demonstrate their effectiveness. 
We fellow CasMVSNet~\cite{gu2020cascade} and adopt three level pyramid structure. \textbf{Stage1} indicates that only the first level of the pyramid is used for the SDF network. \textbf{Stage2} means we sample features from the first and second-level volumes and fuse them together. Similarly, \textbf{Stage3} fuses the features from the first, second, and third-level volumes.
As shown in Tab.~\ref{table_ablation_diff_com}, the reconstructed surface quality significantly improves as we increase the cascade stage numbers. Moreover, The introduction of $\mathcal{L}_{pdc}$ and $\mathcal{L}_{pgs}$ further improves the surface quality.
\vspace{-2pt}

\begin{table}
    
	\begin{center}
		\begin{tabular}{ c c c c  c | c}
			\hline
			Stage1 & Stage2 & Stage3 & $\mathcal{L}_{pdc}$ & $\mathcal{L}_{pgs}$ & Cham. Dis \\
			\hline\hline
			$\surd$ & & &   & & 1.42 \\
            \hline
			  & $\surd$ & &  &  & 1.28 \\
                & $\surd$ & & $\surd$ &  & 1.24 \\
                & $\surd$ & & $\surd$ & $\surd$ & 1.18 \\
            \hline
			  &  & $\surd$ &  & & 1.19 \\
                &  & $\surd$ & $\surd$ & $\surd$ & \textbf{1.11} \\
			\hline
		\end{tabular}
	\end{center}
	\vspace{-10pt}
    \caption{Effect of different components. The performance continues to improve as components increase, which demonstrates the effectiveness of each component.}
    \vspace{-10pt}
	\label{table_ablation_diff_com}
\end{table}

\paragraph{Effect of Cross-view and Cross-level Fusion Strategy.}

\begin{table}   
    
	\begin{center}
		\begin{tabular}{ c  | c}
			\hline
			Method &  Cham. Dis \\
			\hline\hline
            Stage3 w/o fusion strategy & 2.71 \\
            \hline
			Stage3 w/ fusion strategy & \textbf{1.19} \\
			\hline
		\end{tabular}
	\end{center}
	\vspace{-10pt}
    \caption{Effect of cross-view and cross-level fusion strategy. The performance severely degrades without the fusion strategy.}
    \vspace{-10pt}
	\label{table_ablation_fusion}
\end{table}

\begin{figure}
\centering
\includegraphics[width=1.\linewidth]{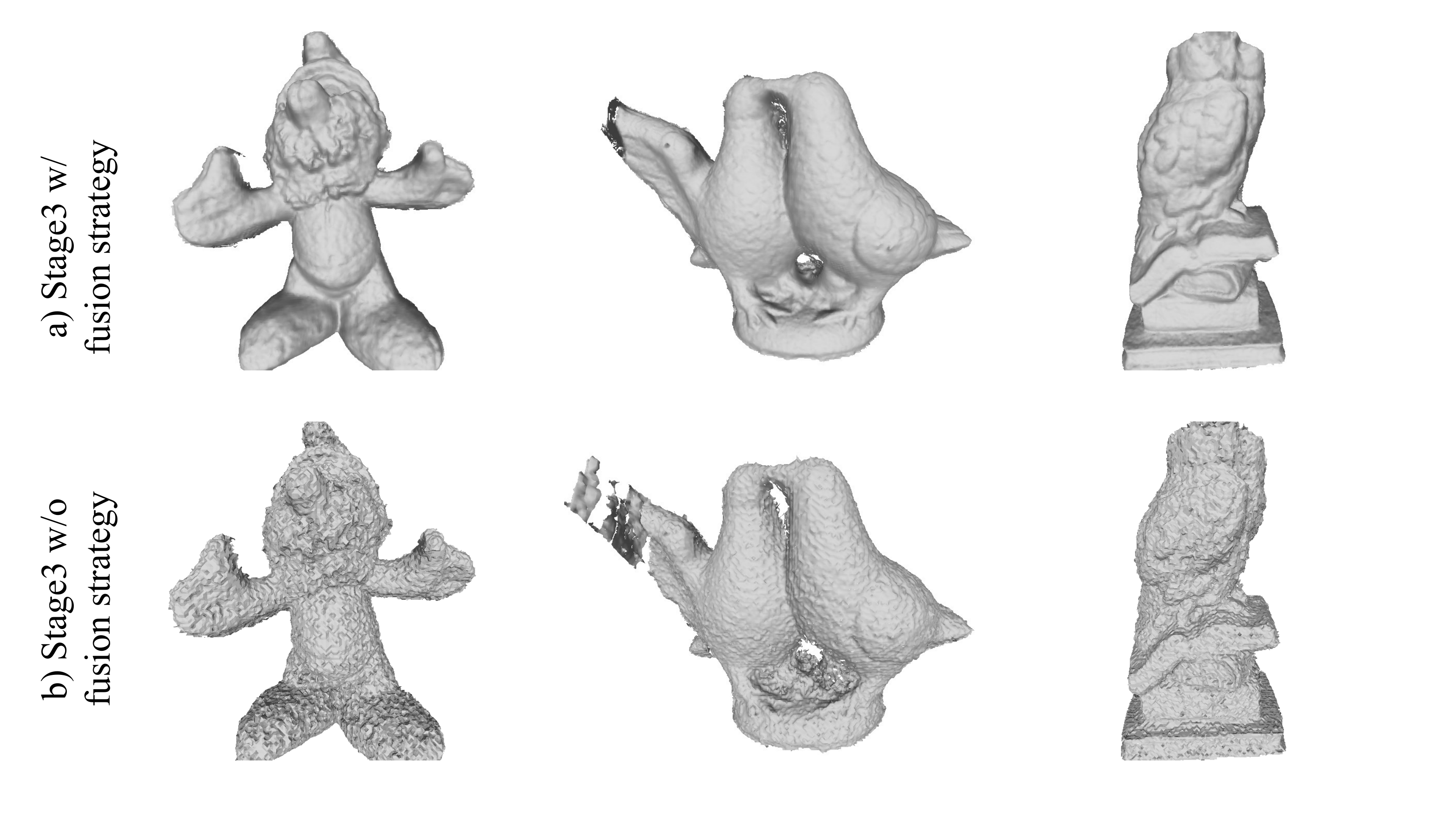}
\vspace{-20pt}
\caption{The qualitative comparison between with and without fusion strategy. Without our fusion strategy, the reconstructed surface becomes noisy.}
\vspace{-10pt}
\label{figure_result_fus}
    
\end{figure}

To demonstrate the effectiveness of the proposed cross-view and cross-level fusion strategy, we remove this strategy on \emph{Stage3} and adopt simple addition. As shown in Tab.~\ref{table_ablation_fusion} and Fig.~\ref{figure_result_fus},  using only simple addition will lead to severe performance degradation and extract noisy surfaces. On the other hand, with our fusion strategy, the performance is significantly improved, and a finer surface is extracted. This demonstrates the effectiveness of our proposed fusion strategy in capturing global-local information and structural consistency.

\section{CONCLUSIONS}
We propose a novel integration scheme, C2F2NeUS, for exploiting both the strengths of MVS and neural implicit surface reconstruction. Previous methods rely on regular euclidean volume for cross-view fusion, which doesn’t simultaneously fit all cameras well and may lead to blurred features. We instead present a cascade geometric frustum for each view and conduct an effective fusion of all the views. Our method achieves state-of-the-art reconstruction quality for sparse inputs, which demonstrates its effectiveness. However, our method still suffers from several limitations, one is that the frustums can overlap with each other in 3D space resulting in redundant computations, and our approach constructs a cost frustum for each view, making it infeasible for dense views. In the future, we plan to optimize the frustum space and reduce computation in overlapping areas.


\textbf{Acknowledgements.}  This work is supported by the National Key R\&D Program of China (No: 2021YFF0500302), the National Natural Science Foundation of China (NSFC No. 62272184), and CCF-Tencent Open Research Fund (CCF-Tencent RAGR20220120). The computation is completed in the HPC Platform of Huazhong University of Science and Technology.

{\small
\bibliographystyle{ieee_fullname}
\bibliography{egbib}
}

\end{document}